\pdfoutput=1

\documentclass[11pt]{article}

\usepackage[preprint]{acl}

\usepackage{times}
\usepackage{latexsym}
\usepackage{amsmath}

\usepackage[T1]{fontenc}

\usepackage[utf8]{inputenc}

\usepackage{microtype}

\usepackage{inconsolata}

\usepackage{graphicx}

\usepackage{mathtools}
\usepackage{makecell}
\usepackage{booktabs}
\usepackage{multirow}
\usepackage{float}
\usepackage{pifont}
\usepackage{tabularx}
\newcommand{\quotes}[1]{``#1''}
\usepackage{amssymb}
\usepackage{subcaption}

\usepackage{tcolorbox}
\tcbuselibrary{listings,skins}
\newtcolorbox{promptbox}[1][]{
  enhanced,
  colback=gray!5,    
  colframe=gray!50,  
  fonttitle=\bfseries,
  coltitle=black,
  title=#1,
  boxrule=0.5pt,
  arc=2mm,           
  outer arc=2mm,
  left=4pt,right=4pt,top=4pt,bottom=4pt,
  after skip=12pt,
  listing only,
  listing options={
    basicstyle=\ttfamily\small,
    breaklines=true,
    columns=fullflexible
  },
  width=0.7\linewidth
}

\title{Mitigating Label Length Bias in Large Language Models}

\author{
    Mario Sanz-Guerrero$^1$ \and Katharina von der Wense$^{1,2}$ \\
    $^1$Johannes Gutenberg University Mainz, Germany \\
    $^2$University of Colorado Boulder, USA \\
    \texttt{\{\href{mailto:msanz@uni-mainz.de}{msanz}, \href{mailto:k.vonderwense@uni-mainz.de}{k.vonderwense}\}@uni-mainz.de}
}

\begin{document}
\maketitle
\begin{abstract}
Large language models (LLMs) are powerful zero- and few-shot learners.
However, when predicting over a set of candidate options, LLMs suffer from label biases, and existing calibration methods overlook biases arising from multi-token class labels. We tackle an issue we call \emph{label length bias}, where labels of different lengths are treated inconsistently, even after standard length normalization.
To mitigate it, we propose \emph{normalized contextual calibration} (NCC), an effective method that normalizes and calibrates predictions at the full-label level.
NCC achieves statistically significant improvements over prior approaches across multiple datasets and models, with gains of up to 10\% F1.
Moreover, NCC extends bias mitigation to broader tasks such as multiple-choice question answering.
Our analysis shows that, when combined with in-context learning, NCC is less sensitive to few-shot example selection, requires fewer examples for competitive performance, and produces more reliable confidence estimates.
These findings highlight the importance of mitigating full-label biases to improve the performance and robustness of LLM-based methods, particularly in real-world applications where class labels naturally consist of multiple tokens.
\end{abstract}

\section{Introduction}

Large language models (LLMs) have demonstrated strong zero- and few-shot capabilities, enabling them to perform new tasks with little or no task-specific supervision. One of the most common ways to leverage these capabilities is through in-context learning (ICL), where models are prompted with a small set of labeled examples to guide their predictions \citep{brown2020fewshot}.

A key challenge when using LLMs to predict over a set of candidate options (e.g., for text classification tasks) is label bias,
where certain labels are inherently favored due to their lexical properties or prior likelihoods: consider a sentiment analysis task where a model must decide between the two classes \quotes{\textit{good}} (the positive class) and \quotes{\textit{atrocious}} (the negative class) -- the model would be biased toward generating \quotes{\textit{good}}, as it has seen this word more frequently in its training data. \citet{zhao2021calibrate} highlights the need for calibration to ensure predictions reflect input alignment rather than model-internal biases. However, existing calibration techniques are insufficient for multi-token class labels, as they typically consider only single-token labels. This approach ignores the joint probability of full-label sequences, which are frequently found in practical scenarios.

\begin{figure}
    \centering
    \includegraphics[width=\linewidth]{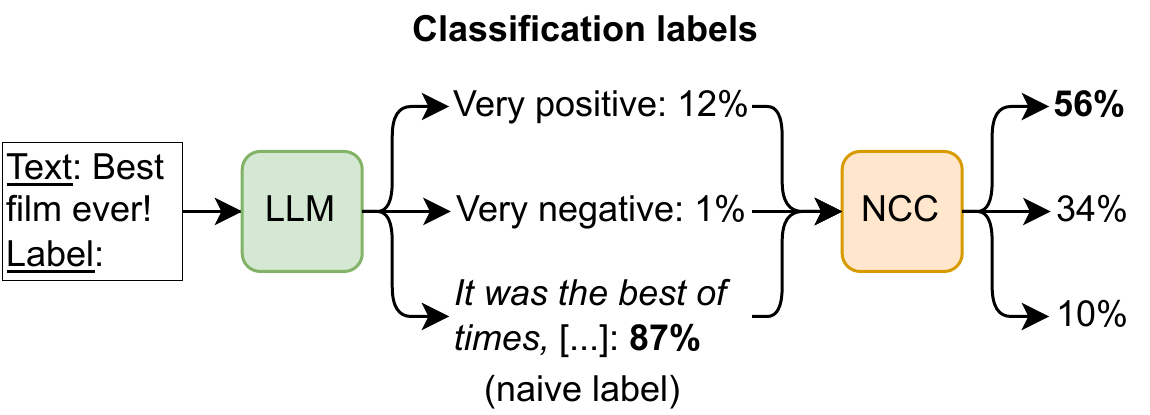}
    \caption{NCC enables the use of calibration in multi-token tasks, mitigating label biases of LLMs and improving their performance. The figure shows real numbers obtained with the Llama 3.1 (8B) model.}
    \label{fig:ncc_diagram}
\end{figure}

In this work, \textbf{we spotlight a previously overlooked type of bias in LLMs: \emph{label length bias}}, where models are biased toward certain classes based on label lengths rather than on semantic alignment with the input.
\textbf{To mitigate this bias, we propose \emph{normalized contextual calibration} (NCC)}, a novel calibration method that accounts for full-label probabilities. NCC first normalizes each label's multi-token probability by its length. It then calibrates the model's predictions by dividing each normalized probability by the corresponding prior probability -- computed from content-free inputs -- to explicitly adjust for label biases.

We evaluate NCC across multiple datasets and models. Our results show that \textbf{NCC consistently outperforms previous approaches}, achieving statistically significant improvements of up to 10\% F1.
Moreover, \textbf{NCC extends beyond text classification, proving effective in multiple-choice QA}, highlighting its adaptability to different NLP tasks. Additionally, \textbf{our method shows to be more robust to few-shot example selection and order, requires fewer examples to achieve competitive performance, and the confidences of its predictions are more reliable} than prior methods.
By addressing the challenge of multi-token label calibration, NCC enhances the applicability of LLM-based methods to real-world settings where class labels often consist of multiple tokens.

\section{Related Work}

\paragraph{In-context Learning}
Several studies have explored various factors influencing ICL's effectiveness. For instance, \citet{lu2022fantastically} demonstrates that ICL is highly sensitive to the order of in-context examples, while \citet{liu2022whatmakes} shows that selecting examples semantically similar to the input can enhance performance. \citet{min2022rethinking} finds that the prompt structure, rather than the specific label values, is a critical factor in classification tasks using ICL.
Recent work has shown that ICL is susceptible to \emph{input} length bias, learning to associate labels with inputs based on the input lengths rather than their semantic correctness \cite{schoch2025inputlengthbias}.
Complementary to these works, we focus on \emph{label} length biases and their mitigation.

\paragraph{Label Bias Mitigation}
\citet{zhao2021calibrate} introduces contextual calibration (CC), which uses a content-free input (e.g., \quotes{N/A} or the empty string) to estimate the biased distribution of label probabilities. This method assumes that the label probabilities for a context-free input should ideally be uniform. The resulting biases are then used to calibrate the model's predictions. The authors also categorize biases in ICL into three types: \emph{majority bias} (favoring frequent labels in the prompt), \emph{recency bias} (favoring recently seen examples), and \emph{common token bias} (favoring commonly occurring tokens in the pretraining distribution). However, their approach primarily addresses datasets with few and short tokenized labels, often assuming that class labels are composed of a single token. For multi-token labels, they only consider the first token (e.g., \quotes{\texttt{Ab}} for the \quotes{\textit{Abbreviation}} label in the TREC dataset \citep{voorhees2000trec}), overlooking the influence of the full label. 

Building on this idea, \citet{fei2023domaincalibration} highlights a new type of label bias (\emph{domain-label bias}) and proposes domain-context calibration to mitigate it. This method adjusts for domain-specific jargon but still relies on single-token labels. Other recent calibration techniques similarly focus on single-token labels \cite{han2023prototypicalcalibration,jiang2023generative,zhou2024batchcalibration}. To circumvent multi-token issues, authors modify datasets by reducing labels to single tokens, often avoiding subword tokenization through lowercasing.

While such simplifications may work for some datasets, they are not universally feasible. Datasets like SST-5 \cite{socher2013sst2_5} contain multi-token labels such as \quotes{\textit{very positive}} and \quotes{\textit{very negative}}, which become indistinguishable if only the first token (\quotes{\texttt{very}}) is used. Prior works replace such labels with single-token alternatives (e.g., \quotes{\textit{very positive}} with \quotes{\textit{great}}, and \quotes{\textit{very negative}} with \quotes{\textit{terrible}}). However, this approach fails for datasets with many or semantically rich labels
(e.g., nuances in labels like \quotes{\textit{card arrival}} and \quotes{\textit{card delivery estimate}} from the Banking77 dataset \citep{banking77_dataset} cannot be adequately represented by a single token).
Moreover, representing multi-token labels only by their first token overgeneralizes, as this token may carry probability mass associated with unrelated labels (e.g., \quotes{\texttt{Ab}} in TREC could correspond to \quotes{\textit{\underline{Ab}ility}} or \quotes{\textit{\underline{Ab}out}}).

To address these limitations, \citet{milios2023manylabels} proposes a method in which the model freely generates a label -- potentially one that does not exist in the predefined set -- and then matches it to the most similar label using a sentence similarity model (e.g., SBERT \citep{reimers2019sbert}). While this avoids the tokenization issue and enables the use of LLMs for text classification with many (and multi-token) labels, it introduces potential errors from the similarity model, which may distort the evaluation of the LLM. Furthermore, calibration cannot be applied to the predictions, as this method relies on free-text generation rather than label probabilities.

Motivated by the absence of a standardized method that fully accounts for multi-token class labels, we first analyze a previously overlooked type of label bias, which we term \emph{label length bias}. To address this, we propose NCC, a label bias mitigation method that explicitly considers the entire label during calibration.

\section{Label Length Bias in LLMs}
\label{sec:length_bias}

\begin{figure}
    \centering
    \includegraphics[width=\linewidth]{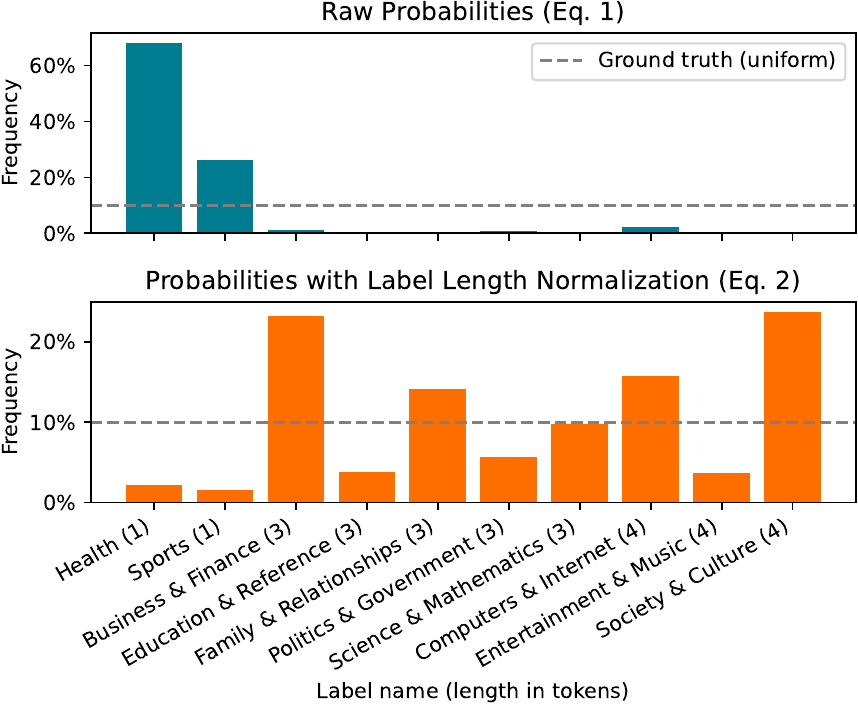}
    \caption{Prediction frequency by label using zero-shot Llama 3.1 (8B) in the Yahoo dataset \citep{agnews_dbpedia_yahoo_dataset}.}
    \label{fig:label_length_bias}
\end{figure}

\paragraph{LLMs Are Susceptible to Length Bias}
LLMs are widely used for classification tasks by predicting over a set of candidate labels, either in few-shot (e.g., via ICL) or zero-shot settings. Let $\mathcal{L}$ be the set of possible class labels, where each label $y \in \mathcal{L}$ consists of one or more tokens. Given an input $x$ (and optionally a context $C_k$ of $k$ labeled examples), the model $M$ predicts the label by computing $\operatorname{argmax}_{y \in \mathcal{L}} P_M(y \mid C_k, x)$. In practice, the probability of a multi-token label $y$ is computed as the product of the conditional probabilities of its tokens \citep{bengio2003neural, shannon1948probs}:
\begin{equation} \label{eq:raw_probs}
    P_M(y \mid C_k, x) = \prod_{i=1}^{n} P_M(t_i \mid C_k, x, t_1, \ldots, t_{i-1})
\end{equation}
where $y$ consists of $n$ tokens $t_1, t_2, \ldots, t_n$.
However, this formulation inherently penalizes longer labels, as each additional token multiplies the probability by a number less than one. As a result, longer labels tend to receive lower overall probabilities than shorter ones, regardless of their semantic appropriateness for the input.
The top plot in Figure~\ref{fig:label_length_bias} demonstrates this phenomenon: without adjusting for length, the model's output probabilities are biased toward shorter labels (e.g., \quotes{\textit{Health}} and \quotes{\textit{Sports}}), which are heavily over-predicted.
We denote this as \emph{label length bias}, a newly-identified type of bias in LLM-based classification.
Previous calibration works \cite[e.g.,][]{zhao2021calibrate,fei2023domaincalibration} overlook this type of bias by focusing only on single-token labels and, thus, fail to address it effectively.

\paragraph{Length Normalization Is Not Enough}
One plausible approach to solve the label length bias in LLMs is to normalize multi-token probabilities by label length \citep{murray2018normalization,brown2020fewshot}. By averaging token probabilities, longer labels are no longer penalized purely for their length.
However, this introduces a different artifact:
in multi-token labels, the probability of the first token is often much lower than that of subsequent tokens, since later tokens are conditioned on the earlier ones and can become highly predictable (e.g., after generating \quotes{\textit{Business~\&}}, the token \quotes{\textit{Finance}} may have near-certain probability\footnote{Even more so if a single word gets splitted into several tokens (e.g., \quotes{\textit{Computers}} \textrightarrow \quotes{\textit{Compu}} + \quotes{\textit{ters}}).}).
This conditioning inflates the normalized score of frequent or highly predictable label sequences,
biasing predictions toward such labels regardless of their overall semantic alignment with the input.
As shown in the bottom plot of Figure~\ref{fig:label_length_bias}, predictions are now better distributed across label lengths, but this leads to over-prediction of labels with likely token continuations (e.g., \quotes{\textit{Business \& Finance}}).
Therefore, without further correction, classification is skewed toward token sequences that the model expects to see, rather than labels that truly align with the input context.

\section{Normalized Contextual Calibration}
In this section, we present NCC, a label bias mitigation method which accounts for label length bias.

When computing the probabilities of multi-token class labels in LLMs, it is essential to address the impact of label length bias. To mitigate it, we propose normalizing the probabilities of labels by their token count. Instead of the raw product of token probabilities, we calculate a normalized probability that accounts for the label length $n$ \citep{jurafsky2000speech, murray2018normalization}:
\begin{equation}\label{eq:normalization}
    P_M^\text{norm}(y \mid C_k, x) = \sqrt[n]{P_M(y \mid C_k, x)}
\end{equation}
This normalization transforms the raw probability into a geometric mean, balancing the contribution of each token and removing the inherent bias against longer labels in LLMs.
While there are different ways of normalizing the label length (e.g., average cross-entropy), we use the geometric mean of token probabilities because it allows for a direct application of calibration (where all labels must sum to 1) and is easier to interpret, allowing for confidence reliability analysis (Section \ref{sec:analysis}).

Once we have comparable label probabilities unbiased by length, we need to mitigate intrinsic label biases. Following prior works \cite{zhao2021calibrate, fei2023domaincalibration}, we use a \emph{content-free} input $x_\varnothing$ -- such as an empty string or a neutral placeholder -- to probe the model and measure the baseline probability it assigns to each label in the absence of meaningful context. We then calibrate the model's output probabilities for input $x$ as follows:
\begin{equation}\label{eq:calibration}
    P_M^\text{calibrated}(y \mid C_k, x) = \frac{P_M^\text{norm}(y \mid C_k, x)}{P_M^\text{baseline}(y \mid C_k, x_\varnothing)}
\end{equation}

The final predicted label with NCC is obtained by selecting the label with the highest calibrated probability:
\begin{equation}
    \hat{y}_{\textrm{NCC}} = \underset{y \in \mathcal{L}}{\operatorname{argmax}}\ P_M^\text{calibrated}(y \mid C_k, x)
\end{equation}

In summary, while normalization addresses biases related to label length, calibration is crucial to ensure that the model's predictions are driven by the input context, rather than by prior biases toward well-known or frequent label sequences.

\section{Experimental Setup}

\paragraph{Datasets}
We evaluate NCC on eight text classification datasets that feature multi-token class labels; cf. Table \ref{tab:token_counts}
(and Appendix \ref{app:dataset} for details). These datasets are widely used in prior studies on ICL \citep{zhao2021calibrate, min2022rethinking, fei2023domaincalibration, milios2023manylabels, bertsch2024longcontext}. Table \ref{tab:token_counts}
confirms that all datasets contain multi-token labels for all models.

\begin{table}
    \centering
    \small
    \setlength{\tabcolsep}{2pt}
    \begin{tabular}{lcccc}
        \toprule
        Dataset & Llama 3.1 & Mistral 7B & Qwen2.5 & GPT-J \\
        \midrule
        AG News & $1.50_{0.87}$ & $1.50_{0.87}$ & $1.50_{0.87}$ & $1.50_{0.87}$ \\ 
        SST-5 & $1.40_{0.49}$ & $1.40_{0.49}$ & $1.40_{0.49}$ & $1.40_{0.49}$ \\ 
        Yahoo & $2.90_{1.04}$ & $3.00_{1.18}$ & $2.90_{1.04}$ & $3.00_{1.10}$ \\ 
        DBpedia & $1.64_{0.72}$ & $1.86_{1.12}$ & $1.64_{0.72}$ & $1.93_{0.88}$ \\ 
        20 Newsgroups & $2.75_{1.22}$ & $3.10_{1.55}$ & $2.75_{1.22}$ & $2.80_{1.29}$ \\ 
        TREC-50 & $2.52_{1.62}$ & $2.80_{1.67}$ & $2.52_{1.62}$ & $2.74_{1.66}$ \\ 
        Banking77 & $3.51_{1.21}$ & $3.57_{1.38}$ & $3.51_{1.21}$ & $3.75_{1.26}$ \\ 
        CLINC150 & $2.09_{0.83}$ & $2.15_{0.95}$ & $2.10_{0.89}$ & $2.39_{0.89}$ \\ 
        \bottomrule
    \end{tabular}
    \caption{Mean and standard deviation of label token lengths for each dataset and model.}
    \label{tab:token_counts}
\end{table}

\paragraph{Models}
We evaluate NCC on four LLMs from different families and of different sizes: Llama 3.1 (8B) \citep{grattafiori2024llama3}, Mistral 7B v0.3 \citep{jiang2023mistral7b}, Qwen2.5 (7B) \citep{qwen2024qwen25}, and GPT-J (6B) \citep{wang2021gptj}. These models also differ in how they tokenize class labels (Table \ref{tab:token_counts}). In order to be able to experiment with a large set of models at a reasonable computational cost, we mostly experiment with LLMs with 6B--8B parameters. However, to demonstrate that NCC is also effective for larger models, we additionally evaluate it on the Llama 3.1 70B version \citep{grattafiori2024llama3}.

\paragraph{Implementation Details}
Following prior work \citep{min2022rethinking, fei2023domaincalibration}, we employ simple and unified prompt templates for all datasets, avoiding task-specific instructions (e.g., explicit label lists) to minimize human engineering.
We use $k = 5$ few-shot examples to enable the models to identify task-specific patterns, which are selected randomly, picking an equal number of examples per class as much as possible. Each experiment is repeated five times with different random seeds to account for the sensitivity of ICL to example selection \citep{liu2022whatmakes}. Additionally, we evaluate NCC in the zero-shot setting ($k = 0$) to analyze its robustness without task demonstrations.
The prompt formats are described in Appendix \ref{app:prompts}.
Inference is performed using NVIDIA A100 GPUs.

To calibrate model biases (Eq.~\ref{eq:calibration}), we use an ensemble of content-free inputs, as proposed by \citet{zhao2021calibrate}. Specifically, we compute the probabilities for each label using five semantically neutral inputs -- \texttt{``''} (empty string), \texttt{``~''} (single space), \texttt{``N/A''}, \texttt{``[MASK]''}, and \texttt{``Lorem ipsum''} -- and average them to reduce variance.

\paragraph{Evaluation}
We report the macro-F1, which accounts for class imbalance (see Table \ref{tab:full_datasets}). To assess statistical significance, we apply the Wilcoxon signed-rank test \cite{wilcoxon1945test}.

\paragraph{Baselines}
We compare NCC against four baselines:
(1) standard raw probabilities (Raw Prob; Eq.~\ref{eq:raw_probs}),
(2) raw probabilities with length normalization (Norm Prob; Eq.~\ref{eq:normalization}),
(3) contextual calibration \citep[CC;][]{zhao2021calibrate},\footnote{We adopt CC as a representative standard calibration baseline due to its simplicity and wide recognition in the literature. For completeness, we compare NCC against other relevant calibration techniques in Appendix~\ref{app:other_calibration_methods}.}\textsuperscript{,}\footnote{For CC, we extend the original single-token implementation in \citet{zhao2021calibrate} with Eq.~\ref{eq:calibration}, where the baseline probabilities are obtained from Eq.~\ref{eq:raw_probs} -- with the single-token implementation, labels ``\textit{very positive}'' and ``\textit{very negative}'' would be treated as the same token (``\texttt{very}'').} and
(4) Gen+SBERT \citep{milios2023manylabels}, a retrieval-based ICL approach that first retrieves the $k$ most similar few-shot examples, then lets the LLM freely generate a label (not restricted to the label space), and finally matches the generated label to the closest candidate using a sentence similarity model (e.g., SBERT).

\section{Results}
\label{sec:results}

\begin{table*}[h]
    \small
    \centering
    \setlength{\tabcolsep}{4pt}
    \begin{tabular}{lcccccccc|c}
    \toprule
    Method & AG News & SST-5 & Yahoo & DBpedia & 20 Newsgroups & TREC-50 & Banking77 & CLINC150 & Avg. \\
    \midrule
    \multicolumn{10}{c}{\textit{Llama 3.1 (8B)}} \\
    \midrule
    Raw Prob & $83.8_{6.2}$ & $\mathbf{42.3_{9.1}^*}$ & $61.3_{2.7}$ & $83.4_{0.7}$ & $41.0_{3.4}$ & $28.2_{1.9}$ & $37.8_{4.6}$ & $55.9_{1.2}$ & $54.2_{3.7}$ \\
    Norm Prob & $\mathbf{85.3_{1.8}^*}$ & $38.6_{6.3}$ & $60.6_{2.2}$ & $80.0_{4.7}$ & $43.3_{3.2}$ & $29.3_{2.4}$ & $36.3_{3.5}$ & $64.1_{2.2}$ & $54.7_{3.3}$ \\
    CC & $84.0_{2.3}$ & $23.9_{7.4}$ & $8.1_{3.5}$ & $1.0_{0.0}$ & $5.3_{6.1}$ & $0.2_{0.0}$ & $0.1_{0.0}$ & $1.0_{1.1}$ & $15.4_{2.6}$ \\
    Gen+SBERT & $80.4_{5.1}$ & $41.2_{5.3}$ & $64.3_{1.1}$ & $67.3_{3.9}$ & $40.5_{2.1}$ & $\mathbf{33.8_{2.9}}$ & $47.5_{3.1}$ & $52.3_{3.2}$ & $53.4_{3.3}$ \\
    \cmidrule(lr){2-10}
    NCC & $78.0_{5.0}$ & $38.2_{8.1}$ & $\mathbf{65.1_{3.5}}$ & $\mathbf{92.5_{0.7}^*}$ & $\mathbf{67.7_{0.9}^*}$ & $\mathbf{33.8_{2.1}}$ & $\mathbf{59.5_{0.9}^*}$ & $\mathbf{73.1_{1.0}^*}$ & $\mathbf{63.5_{2.8}^*}$ \\
    \midrule
    \multicolumn{10}{c}{\textit{Mistral 7B}} \\
    \midrule
    Raw Prob & $81.2_{6.4}$ & $40.7_{7.7}$ & $52.9_{3.2}$ & $67.7_{5.8}$ & $39.1_{4.1}$ & $28.6_{1.4}$ & $34.5_{3.9}$ & $53.1_{1.6}$ & $49.7_{4.3}$ \\
    Norm Prob & $\mathbf{83.1_{2.8}^*}$ & $33.4_{6.1}$ & $56.2_{3.2}$ & $62.6_{3.1}$ & $32.5_{7.1}$ & $26.8_{5.3}$ & $32.9_{5.0}$ & $61.9_{2.6}$ & $48.7_{4.4}$ \\
    CC & $81.5_{1.7}$ & $21.1_{3.8}$ & $7.4_{3.7}$ & $1.0_{0.0}$ & $4.0_{3.5}$ & $0.2_{0.0}$ & $0.1_{0.0}$ & $1.4_{0.9}$ & $14.6_{1.7}$ \\
    Gen+SBERT & $81.9_{2.3}$ & $\mathbf{40.9_{7.3}^*}$ & $\mathbf{64.3_{1.4}}$ & $62.0_{3.3}$ & $39.2_{3.1}$ & $\mathbf{36.4_{5.1}}$ & $46.9_{1.5}$ & $49.1_{3.1}$ & $52.6_{3.4}$ \\
    \cmidrule(lr){2-10}
    NCC & $76.8_{4.1}$ & $34.7_{7.3}$ & $62.1_{2.0}$ & $\mathbf{91.2_{1.6}^*}$ & $\mathbf{60.4_{4.2}^*}$ & $34.2_{1.8}$ & $\mathbf{57.9_{1.1}^*}$ & $\mathbf{64.8_{1.9}^*}$ & $\mathbf{60.3_{3.0}^*}$ \\
    \midrule
    \multicolumn{10}{c}{\textit{Qwen2.5 (7B)}} \\
    \midrule
    Raw Prob & $74.6_{4.7}$ & $35.8_{4.6}$ & $47.1_{1.5}$ & $74.1_{3.7}$ & $29.4_{4.7}$ & $25.1_{2.4}$ & $23.6_{5.5}$ & $55.6_{2.0}$ & $45.7_{3.6}$ \\
    Norm Prob & $\mathbf{83.1_{2.4}^*}$ & $\mathbf{42.5_{4.3}^*}$ & $44.6_{2.2}$ & $72.8_{4.2}$ & $21.3_{7.5}$ & $19.1_{5.2}$ & $18.3_{4.6}$ & $63.8_{3.4}$ & $45.7_{4.2}$ \\
    CC & $49.2_{5.9}$ & $24.0_{5.9}$ & $35.4_{11.4}$ & $3.1_{1.3}$ & $10.4_{3.6}$ & $0.2_{0.0}$ & $0.1_{0.0}$ & $2.7_{1.4}$ & $15.6_{3.7}$ \\
    Gen+SBERT & $74.6_{3.3}$ & $37.6_{4.8}$ & $\mathbf{59.1_{1.8}}$ & $71.1_{4.2}$ & $30.8_{4.3}$ & $33.8_{5.1}$ & $41.1_{4.1}$ & $54.5_{3.4}$ & $50.3_{3.9}$ \\
    \cmidrule(lr){2-10}
    NCC & $71.7_{3.8}$ & $35.5_{5.8}$ & $\mathbf{59.1_{2.5}}$ & $\mathbf{87.1_{1.9}^*}$ & $\mathbf{58.3_{2.8}^*}$ & $\mathbf{34.6_{2.9}}$ & $\mathbf{52.3_{2.4}^*}$ & $\mathbf{71.1_{2.4}^*}$ & $\mathbf{58.7_{3.1}^*}$ \\
    \midrule
    \multicolumn{10}{c}{\textit{GPT-J (6B)}} \\
    \midrule
    Raw Prob & $67.3_{1.5}$ & $\mathbf{25.7_{3.9}}$ & $30.6_{12.5}$ & $70.8_{3.1}$ & $17.2_{9.6}$ & $24.0_{1.7}$ & $29.8_{2.7}$ & $44.5_{0.9}$ & $38.7_{4.5}$ \\
    Norm Prob & $\mathbf{74.8_{3.2}^*}$ & $25.4_{4.6}$ & $43.5_{14.8}$ & $65.0_{6.2}$ & $10.3_{9.4}$ & $14.5_{5.3}$ & $32.4_{2.3}$ & $51.0_{1.5}$ & $39.6_{5.9}$ \\
    CC & $70.7_{18.0}$ & $12.5_{6.9}$ & $12.4_{4.8}$ & $1.0_{0.0}$ & $4.7_{4.5}$ & $0.2_{0.0}$ & $0.1_{0.0}$ & $1.4_{0.6}$ & $12.9_{4.3}$ \\
    Gen+SBERT & $66.0_{0.7}$ & $25.0_{2.0}$ & $46.4_{14.8}$ & $63.5_{1.5}$ & $18.0_{9.4}$ & $\mathbf{28.2_{3.0}^*}$ & $43.1_{1.0}$ & $48.6_{1.7}$ & $42.3_{4.3}$ \\
    \cmidrule(lr){2-10}
    NCC & $67.7_{1.5}$ & $24.4_{3.5}$ & $\mathbf{46.5_{5.8}}$ & $\mathbf{90.6_{0.8}^*}$ & $\mathbf{47.8_{4.6}^*}$ & $21.7_{1.1}$ & $\mathbf{43.2_{1.6}}$ & $\mathbf{57.4_{3.1}^*}$ & $\mathbf{49.9_{2.8}^*}$ \\
    \midrule
    \multicolumn{10}{c}{\textit{Llama 3.1 (70B)}} \\
    \midrule
    Raw Prob & $\mathbf{88.8_{2.0}^*}$ & $48.0_{8.4}$ & $69.7_{1.3}$ & $80.8_{3.1}$ & $43.2_{4.4}$ & $29.7_{1.2}$ & $36.2_{6.6}$ & $55.8_{2.5}$ & $56.5_{3.7}$ \\
    Norm Prob & $87.9_{1.0}$ & $\mathbf{48.1_{6.2}^*}$ & $\mathbf{70.4_{0.4}^*}$ & $80.1_{2.8}$ & $50.3_{2.9}$ & $26.1_{5.1}$ & $31.0_{6.1}$ & $66.9_{1.9}$ & $57.6_{3.3}$ \\
    CC & $86.5_{1.3}$ & $27.4_{6.8}$ & $53.1_{6.1}$ & $21.5_{10.7}$ & $0.4_{0.0}$ & $0.2_{0.0}$ & $0.8_{0.7}$ & $3.3_{2.1}$ & $24.1_{3.4}$ \\
    Gen+SBERT & $84.3_{2.5}$ & $47.0_{3.4}$ & $62.9_{2.5}$ & $62.5_{3.9}$ & $47.3_{4.7}$ & $\mathbf{37.1_{3.2}^*}$ & $45.9_{5.0}$ & $55.3_{2.8}$ & $55.3_{3.5}$ \\
    \cmidrule(lr){2-10}
    NCC & $84.6_{4.8}$ & $37.8_{3.9}$ & $65.0_{1.3}$ & $\mathbf{98.2_{0.3}^*}$ & $\mathbf{73.5_{1.5}^*}$ & $32.0_{2.8}$ & $\mathbf{56.6_{3.3}^*}$ & $\mathbf{68.5_{2.6}^*}$ & $\mathbf{64.5_{2.6}^*}$ \\
    \bottomrule
    \end{tabular}
    \caption{Few-shot performance (macro-F1$_{\pm\text{std}}$) of the conventional approach (Raw Prob; \citealt{brown2020fewshot}), conventional approach with label length normalization (Norm Prob), contextual calibration (CC; \citealt{zhao2021calibrate}), free generation plus similarity matching (Gen+SBERT; \citealt{milios2023manylabels}), and normalized contextual calibration (NCC; \textit{Ours}). *~indicates significant dominance of (or over) NCC at p~$<$~0.05.}
    \label{tab:results_k5}
\end{table*}

\begin{figure}
    \centering
    \includegraphics[width=\linewidth]{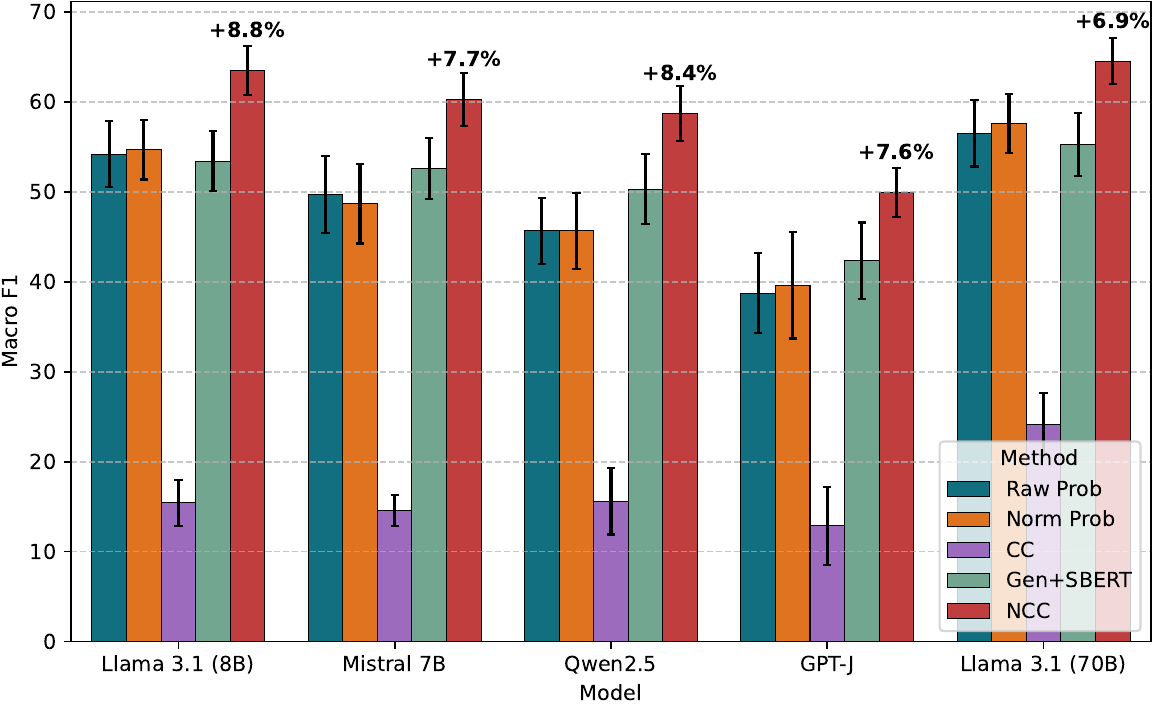}
    \caption{Few-shot performance of methods and models, averaged over all datasets. Error bars indicate the standard deviation across runs. Numbers above the bars indicate absolute improvement of NCC over the second-best method.}
    \label{fig:plot_by_model_and_method_5}
\end{figure}

\paragraph{Few-shot Results}
Table \ref{tab:results_k5} reports detailed results across models, datasets, and methods with $k = 5$ examples, while Figure \ref{fig:plot_by_model_and_method_5} summarizes overall trends. NCC consistently outperforms all baselines with statistically significant improvements, achieving absolute macro-F1 gains ranging from +6.9\% to +8.8\% over the second-best method (7.6\% on average). The gains are most pronounced on datasets with longer and more complex labels, such as Banking77 and CLINC150, where mitigating label length bias is most critical.

NCC also surpasses retrieval-based ICL approaches such as Gen+SBERT, which leverages \emph{similar} few-shot examples and free-form label generation followed by similarity matching to candidate labels.
Despite using \emph{randomly} selected examples, NCC achieves higher performance, suggesting that properly calibrated label probabilities are more beneficial than retrieval similarity alone -- a common approach in practical applications.

Two model-specific trends emerge: (1) NCC consistently improves performance across all model sizes, including Llama 3.1 70B, which shows that even larger LLMs suffer from label bias and require calibration;
(2) NCC narrows the performance gap between large and small models, indicating that proper calibration can make smaller models competitive (see Appendix~\ref{app:model_size} for details).

Notably, the standard CC method for multi-token labels \citep[adapted from][]{zhao2021calibrate} achieves near-zero F1 on some datasets (e.g., TREC-50).
Without length normalization, this method inherits the bias toward shorter labels (higher $P_M^\text{baseline}$). During calibration (Eq.~\ref{eq:calibration}), this results in excessive penalization of short labels and overcompensation for longer ones, skewing predictions toward lengthy labels regardless of input relevance (details in Appendix~\ref{app:reliability}).
This illustrates the necessity of normalizing multi-token probabilities to prevent calibration from amplifying label length bias.

NCC underperforms the standard raw probability on AG News and SST-5, where the full label set (4 and 5 classes, respectively) is already represented in the few-shot examples ($k=5$) and label semantics are simple. In such cases, NCC introduces unnecessary adjustments that degrade performance. This aligns with our zero-shot findings (below), where NCC becomes essential to mitigate label bias when the label space is not fully represented.

\paragraph{Zero-shot Results}\label{subsec:zero_shot}
\begin{figure}[b]
    \centering
    \includegraphics[width=\linewidth]{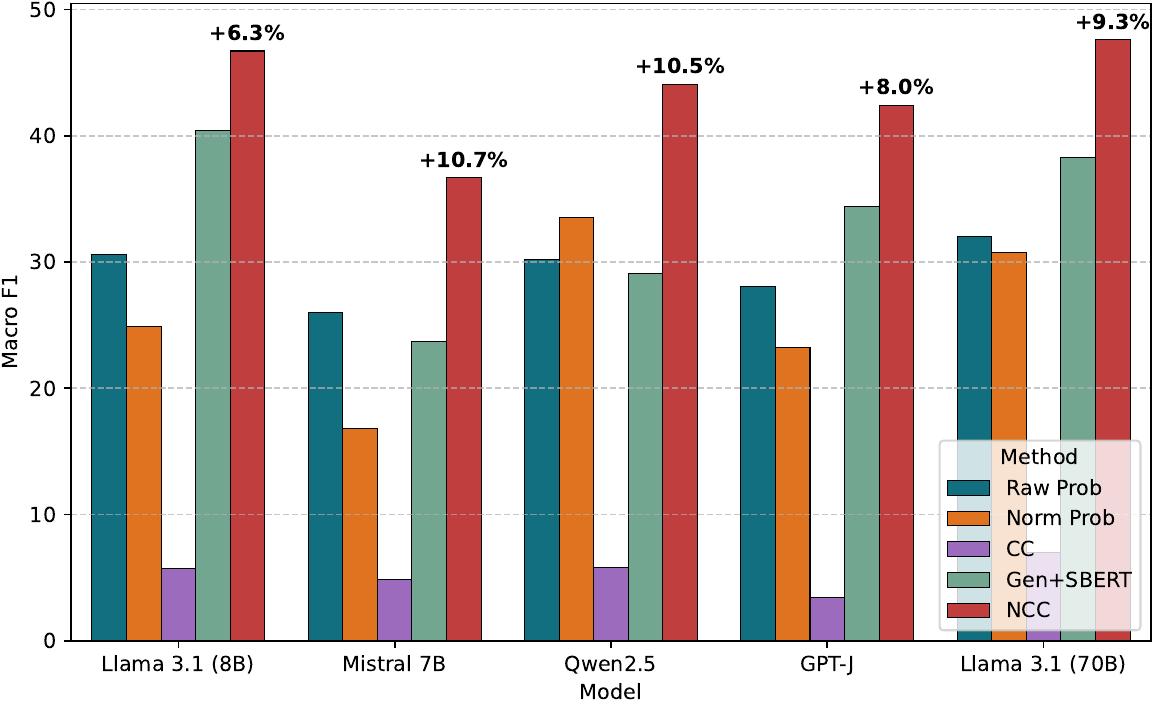}
    \caption{Zero-shot performance of methods and models, averaged over all datasets. Numbers indicate absolute improvement of NCC over the second-best method.}
    \label{fig:plot_by_model_and_method_0}
\end{figure}
\begin{figure}
    \centering
    \includegraphics[width=\linewidth]{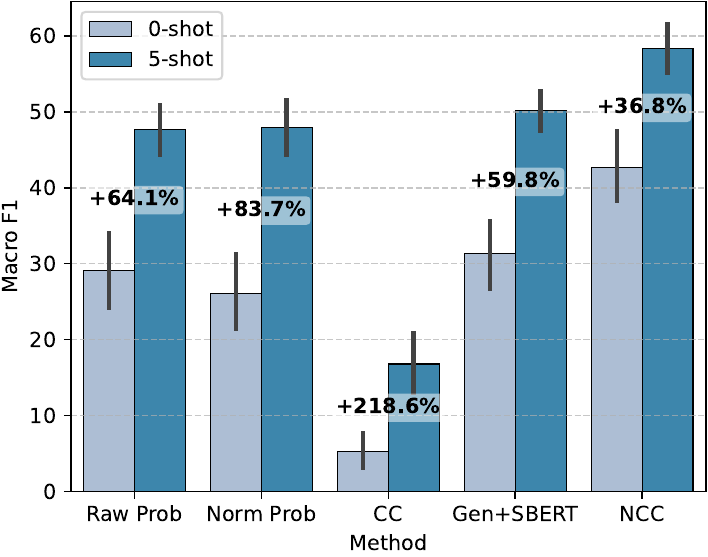}
    \caption{Average performance of all methods in the zero-shot and few-shot setting. Numbers indicate relative improvement when going from 0 to 5 shots.}
    \label{fig:plot_by_method_and_k}
\end{figure}
We further evaluate NCC in a zero-shot setting, where no demonstrations are provided. As shown in Figure~\ref{fig:plot_by_model_and_method_0}, NCC maintains its advantage over all baselines across most datasets and models, mirroring trends from the few-shot scenario (detailed results in Appendix~\ref{app:zero_shot}).

While all methods yield lower performance due to the lack of examples as expected, NCC shows the smallest relative performance difference between zero- and few-shot settings (36.8\%; Figure~\ref{fig:plot_by_method_and_k}), indicating that it is less dependent on labeled examples.
In some cases, NCC's \emph{zero-shot} performance even exceeds the \emph{few-shot} raw baseline (e.g., GPT-J average F1: 42.4 vs.\ 38.7).
This further demonstrates the effectiveness of NCC's label bias mitigation, as it allows models to leverage the unbiased label semantics and perform better on zero-shot than on standard few-shot learning from examples.

Our findings align with those in \citet{min2022rethinking}, who emphasize that defining the label space is crucial for ICL performance. In the few-shot experiments where all possible labels are included in the context (e.g., AG News and SST-5 in Table \ref{tab:results_k5}), the model is already well-calibrated by the explicit definition of the label space, so NCC introduces unnecessary adjustments that degrade performance.
In contrast, in the zero-shot setting, when the label space is only implicitly defined, NCC becomes essential to correct the model's inherent biases and improve performance. This highlights NCC's role as a necessary calibration step when the label space is incomplete or underspecified in the prompt.

\section{Generalizability of NCC to Other Tasks}

Another key contribution of NCC is its ability to extend the applicability of calibration techniques beyond classification tasks. Prior calibration methods (e.g., \citet{zhao2021calibrate}) are limited to scenarios with single-token labels. By extending calibration with label length bias mitigation, NCC enables bias correction in previously inaccessible tasks.

We demonstrate this capability through multiple-choice question answering (MCQA), a task that requires selecting the correct option for a question among a set of choices.
For this experiment, we adopt a cloze prompt formulation \cite{brown2020fewshot}, a widely used MCQA approach that involves getting the model probabilities for each independent completion (choice), without showing the list of options to the model (detailed prompt can be found in Appendix \ref{app:prompts}). Thus, the model does not know the label distribution\footnote{In MCQA, the ``labels'' are the different choices.} and might exhibit label biases.
Table \ref{tab:results_qa} evaluates NCC on three diverse MCQA datasets.
The results highlight the effectiveness of NCC in these tasks, achieving the highest average performance across all models.

Unlike text classification, MCQA does not exhibit \emph{majority} or \emph{recency} biases \citep{zhao2021calibrate}, as each example has its own unique set of candidate answers. Instead, the primary challenge lies in addressing \emph{common token} and \emph{label length} biases, which can distort the model's probability estimates.
While gains are smaller than in text classification -- likely due to the absence of \emph{majority} and \emph{recency} biases, diminishing the effect of calibration -- these consistent improvements demonstrate the potential of NCC to generalize beyond text classification. By enabling calibration techniques to handle multi-token labels (or options) effectively, NCC opens the door to applying these methods in a wide array of tasks that require mitigation of label biases.

\begin{table}[t]
    \small
    \centering
    \setlength{\tabcolsep}{4pt}
    \begin{tabular}{lccc|c}
    \toprule
    Method & OBQA & CSQA & QASC & Avg. \\
    \midrule
    \multicolumn{5}{c}{\textit{Llama 3.1 (8B)}} \\
    \midrule
    Raw Prob & $24.8_{0.5}$ & $55.7_{1.4}$ & $51.3_{4.4}$ & $43.9_{2.1}$ \\
    Norm Prob & $32.8_{1.0}$ & $57.3_{1.2}$ & $51.6_{4.1}$ & $47.2_{2.1}$ \\
    CC & $33.2_{1.2}$ & $34.7_{0.8}$ & $22.8_{2.0}$ & $30.2_{1.3}$ \\
    Gen+SBERT & $29.8_{1.2}$ & $45.7_{0.5}$ & $43.9_{1.6}$ & $39.8_{1.1}$ \\
    \cmidrule(lr){2-5}
    NCC & $\mathbf{37.9_{1.0}}$ & $\mathbf{58.4_{0.8}}$ & $\mathbf{59.3_{2.9}}$ & $\mathbf{51.9_{1.6}}$ \\
    \midrule
    \multicolumn{5}{c}{\textit{Mistral 7B}} \\
    \midrule
    Raw Prob & $23.2_{1.2}$ & $57.5_{0.9}$ & $52.8_{2.8}$ & $44.5_{1.6}$ \\
    Norm Prob & $\mathbf{33.7_{1.4}}$ & $56.9_{1.4}$ & $52.2_{2.0}$ & $47.6_{1.6}$ \\
    CC & $33.3_{1.5}$ & $31.1_{0.9}$ & $18.0_{1.2}$ & $27.5_{1.2}$ \\
    Gen+SBERT & $29.3_{1.0}$ & $47.8_{1.0}$ & $44.3_{0.4}$ & $40.5_{0.8}$ \\
    \cmidrule(lr){2-5}
    NCC & $32.1_{1.1}$ & $\mathbf{57.6_{1.2}}$ & $\mathbf{57.4_{1.9}}$ & $\mathbf{49.0_{1.4}}$ \\
    \midrule
    \multicolumn{5}{c}{\textit{Qwen2.5 (7B)}} \\
    \midrule
    Raw Prob & $25.7_{1.0}$ & $59.5_{0.7}$ & $54.5_{3.3}$ & $46.6_{1.7}$ \\
    Norm Prob & $33.2_{0.5}$ & $\mathbf{60.5_{0.4}}$ & $53.0_{1.9}$ & $48.9_{0.9}$ \\
    CC & $33.6_{0.8}$ & $36.4_{1.3}$ & $20.4_{0.9}$ & $30.2_{1.0}$ \\
    Gen+SBERT & $31.4_{1.3}$ & $52.5_{0.4}$ & $46.5_{0.7}$ & $43.4_{0.8}$ \\
    \cmidrule(lr){2-5}
    NCC & $\mathbf{39.5_{0.9}}$ & $57.7_{1.2}$ & $\mathbf{62.6_{1.5}}$ & $\mathbf{53.3_{1.2}}$ \\
    \midrule
    \multicolumn{5}{c}{\textit{GPT-J (6B)}} \\
    \midrule
    Raw Prob & $17.0_{1.1}$ & $45.3_{2.0}$ & $37.8_{2.0}$ & $33.4_{1.7}$ \\
    Norm Prob & $23.0_{1.2}$ & $44.5_{1.3}$ & $38.3_{1.4}$ & $35.3_{1.3}$ \\
    CC & $\mathbf{30.3_{0.9}}$ & $24.8_{0.7}$ & $13.4_{0.4}$ & $22.8_{0.7}$ \\
    Gen+SBERT & $23.9_{0.6}$ & $38.2_{1.4}$ & $38.0_{1.2}$ & $33.3_{1.1}$ \\
    \cmidrule(lr){2-5}
    NCC & $26.3_{1.5}$ & $\mathbf{46.9_{0.6}}$ & $\mathbf{43.5_{1.4}}$ & $\mathbf{38.9_{1.2}}$ \\
    \midrule
    \multicolumn{5}{c}{\textit{Llama 3.1 (70B)}} \\
    \midrule
    Raw Prob & $29.1_{0.5}$ & $61.6_{1.8}$ & $60.0_{1.6}$ & $50.2_{1.3}$ \\
    Norm Prob & $37.4_{1.1}$ & $63.5_{1.9}$ & $58.4_{1.1}$ & $53.1_{1.4}$ \\
    CC & $36.5_{0.7}$ & $37.2_{2.0}$ & $25.4_{0.5}$ & $33.0_{1.1}$ \\
    Gen+SBERT & $33.1_{1.8}$ & $51.2_{1.0}$ & $49.9_{0.9}$ & $44.8_{1.2}$ \\
    \cmidrule(lr){2-5}
    NCC & $\mathbf{44.4_{1.7}}$ & $\mathbf{64.1_{1.8}}$ & $\mathbf{68.1_{0.7}}$ & $\mathbf{58.8_{1.4}}$ \\
    \bottomrule
    \end{tabular}
    \caption{Performance (Macro F1) for the multiple choice QA task. Dataset abbreviations: OBQA (OpenBookQA; \citealt{openbookqa_dataset}), CSQA (CommonsenseQA; \citealt{commonsenseqa_dataset}), QASC (Question Answering in Context; \citealt{qasc_dataset}).}
    \label{tab:results_qa}
\end{table}

\section{Analysis}\label{sec:analysis}

\begin{figure*}
    \centering
    \includegraphics[width=\linewidth]{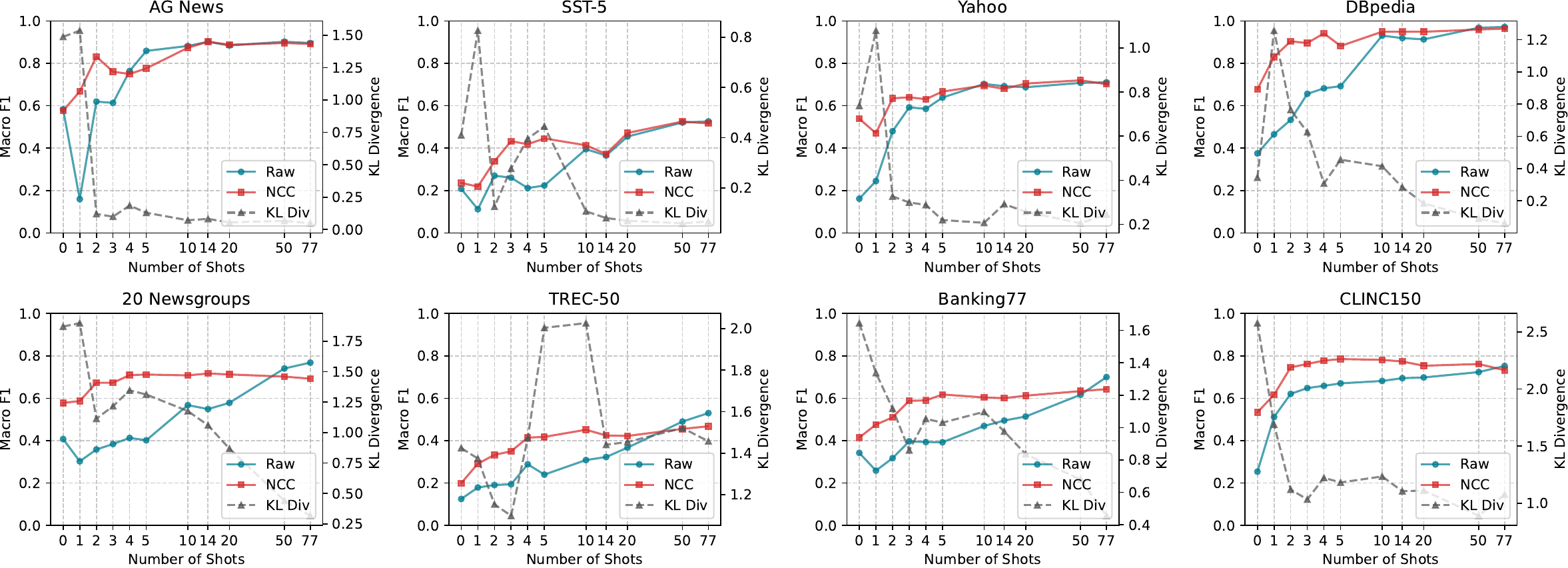}
    \caption{Performance and calibration level for different $k$ values, run with Llama 3.1 (8B).}
    \label{fig:calibration_vs_k}
\end{figure*}

\paragraph{NCC Reduces Sensitivity to Example Selection}
Prior work has shown that ICL is highly sensitive to the choice and order of few-shot examples \citep{liu2022whatmakes, lu2022fantastically}. To assess whether NCC reduces this sensitivity, we compare the average standard deviation and coefficient of variation of F1 scores across different random seeds for each model--dataset combination. 
As shown in Table~\ref{tab:sensitivity}, NCC exhibits the lowest variability, indicating greater robustness and stability across different example sets.

\begin{table}
    \small
    \centering
    \begin{tabular}{lrr}
        \toprule
        Method & Std. Dev. & Coef. Var. \\
        \midrule
        Raw Prob & $0.039$ & $0.103$ \\
        Norm Prob & $0.042$ & $0.134$ \\
        CC & $0.034$ & $0.303$ \\
        Gen+SBERT & $0.038$ & $0.088$ \\
        \cmidrule(lr){2-3}
        NCC & $\mathbf{0.031}$ & $\mathbf{0.058}$ \\
        \bottomrule
    \end{tabular}
    \caption{Average sensitivity (standard deviation and coefficient of variation) of F1 scores across different random example selections.}
    \label{tab:sensitivity}
\end{table}

\paragraph{Calibration Level vs.\ Few-shot Examples}
We analyze how the number of few-shot examples ($k$) affects calibration by comparing NCC and the standard Raw Prob baseline across varying $k$'s (equal to each dataset's class count) on all datasets.
We select examples randomly from the train set, smaller sets of examples are always subsets of larger sets, and we maintain a fixed order to avoid permutation sensitivity \citep{lu2022fantastically}.
We measure the required calibration level using the Kullback-Leibler (KL) divergence
between the raw and the calibrated (NCC) probability distributions, reflecting the divergence between them. Due to computational cost, we use Llama 3.1 (8B) with a single random seed.

Figure \ref{fig:calibration_vs_k} shows that NCC generally outperforms Raw Prob, with the performance gap narrowing as $k$ increases. This trend is expected because fewer examples require higher calibration to correct biases, as the model has fewer data to learn the label space. The decreasing KL divergence with increasing $k$ confirms this, indicating that raw predictions become more aligned with calibrated probabilities. This suggests NCC is particularly effective in scenarios with limited few-shot examples, where calibration is crucial for mitigating label biases. Additionally, in settings with a large number of classes (bottom row of Figure \ref{fig:calibration_vs_k}), very few examples (2--5) in NCC perform almost as good as the many-examples (50--77) raw ICL counterparts. This has significant implications for practical applications, as the number of examples can be drastically reduced while keeping comparable performance.

\begin{figure}
    \centering
    \includegraphics[width=0.6\linewidth]{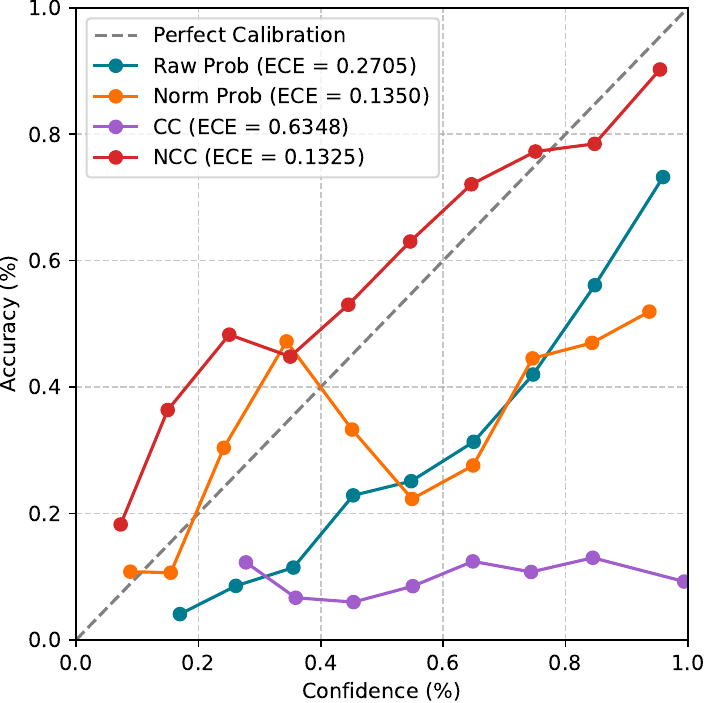}
    \caption{Reliability diagrams for all methods. The x-axis are bins of predicted confidence, and the y-axis shows the accuracy for the given bin.}
    \label{fig:reliability}
\end{figure}

\paragraph{NCC Enhances Reliability of Predictions}
Another crucial aspect when evaluating LLMs is considering how reliable its predictions are. This is measured by the alignment between model confidence (probability) and prediction accuracy \cite{guo2017calibration}. Figure \ref{fig:reliability} shows the reliability diagrams for the different methods, averaged across all datasets for the Llama 3.1 (8B) model. NCC is the most reliable method, showing the closest line to the perfect calibration and the lowest expected calibration error (ECE). This result demonstrates that NCC not only improves overall model performance, but also the reliability of the predictions. (Further analysis is conducted in Appendix \ref{app:reliability}.)

\section{Conclusion}

In this work, we highlight the previously overlooked issue of multi-token label length bias in LLMs and introduce NCC as an effective solution. By normalizing probabilities and applying calibration with prior probabilities at the full-label level, NCC significantly improves performance over existing methods across diverse datasets and models. Our approach extends bias mitigation to settings such as multiple-choice question answering, demonstrating its versatility beyond classification tasks. Additionally, NCC reduces sensitivity to few-shot example selection, requires fewer examples to perform competitively, and produces more reliable confidence estimates. Our findings underscore the importance of addressing full-label biases to enhance the performance and reliability of LLM-based systems, especially in real-world scenarios where class labels often consist of multiple tokens.

\section*{Limitations}  
While NCC demonstrates strong performance and generalizability across various tasks, several limitations should be considered.

First, the applicability of NCC is constrained by the requirement to access token-level probabilities for all possible class labels. This restricts its use to open-source models that can be run locally, as many proprietary or cloud-based models do not provide full token probability distributions. Some APIs (e.g., OpenAI's \texttt{logprobs} parameter) offer token probabilities but are typically limited to the top-20 tokens, making NCC infeasible for tasks requiring complete probability distributions.

Second, NCC is not suitable for open-ended tasks where the label space is not predefined. Since the method relies on calibrating prior biases across a fixed set of labels, it cannot be directly applied to tasks with an unrestricted output space, such as open-ended text generation.

Third, NCC is not designed for full-sentence generation tasks. Calibration is most effective when applied to tasks where the label itself carries minimal semantic meaning (e.g., class labels in classification, or sentence continuations in multiple-choice question answering). In contrast, full-sentence responses inherently carry substantial baseline probability due to their coherence and fluency, regardless of calibration. Applying NCC to meaningful text generation would likely penalize fluent responses rather than correct systematic biases.

Despite these limitations, NCC remains a valuable tool for mitigating label bias and improving LLM performance in settings where class labels are well-defined.

\section*{Ethics Statement}
Our work demonstrates that NCC significantly improves LLM performance for text classification and related NLP tasks by mitigating label biases. However, these improvements do not fully eliminate inherent risks, and LLMs remain susceptible to errors. Therefore, we caution against relying solely on LLMs in critical settings, such as medical advice, without appropriate human oversight and domain-specific validation.

\section*{Acknowledgments}
This work was supported by the Carl Zeiss Foundation through the MAINCE project (grant number P2022-08-009).

\bibliography{refs}

\appendix

\section{Prompt Formats}
\label{app:prompts}

For the prompt formats, we follow previous work and employ simple and unified prompt templates for both text classification and question answering tasks \citep{min2022rethinking,fei2023domaincalibration,sanzguerrero2025cicl,shaier2025asking,saadi2025jguwmt25}.

The prompts end with ``\texttt{Label:}'' (for text classification tasks; Figure \ref{fig:icl_prompt_box}) and ``\texttt{Answer:}'' (for question answering tasks; Figure \ref{fig:qa_prompt_box}), and we extract the model's output probabilities for each candidate label/answer starting from that position.
Following \citet{sanzguerrero2025mindthegap}, the first token of each label/answer is prefixed with a space to align with the model's default tokenization, as this has been shown to improve performance and model calibration.

\begin{figure}[h]
    \centering
    \begin{promptbox}[Text Classification Prompt]
        \ttfamily\small
        Text: \{example\_1\} \\
        Label: \{ground\_truth\_1\} \\[1ex]
        Text: \{example\_2\} \\
        Label: \{ground\_truth\_2\} \\[1ex]
        ... \\[1ex]
        Text: \{example\_$k$\} \\
        Label: \{ground\_truth\_$k$\} \\[1ex]
        Text: \{input\_text\} \\
        Label:
    \end{promptbox}
    \caption{Prompt format for text classification tasks.}
    \label{fig:icl_prompt_box}
\end{figure}

\begin{figure}[h]
    \centering
    \begin{promptbox}[Question Answering Prompt]
        \ttfamily\small
        Question: \{example\_1\} \\
        Answer: \{ground\_truth\_1\} \\[1ex]
        Question: \{example\_2\} \\
        Answer: \{ground\_truth\_2\} \\[1ex]
        ... \\[1ex]
        Question: \{example\_$k$\} \\
        Answer: \{ground\_truth\_$k$\} \\[1ex]
        Question: \{input\_text\} \\
        Answer:
    \end{promptbox}
    \caption{Prompt format for question answering tasks.}
    \label{fig:qa_prompt_box}
\end{figure}

\section{Illustration of the Importance of NCC}
\label{app:calibration}

To illustrate the necessity of NCC for handling multi-token labels, we revisit the toy example illustrated in Figure \ref{fig:ncc_diagram}. Consider a sentiment classification task similar to the SST-5 dataset, where the five class labels are: ``\textit{very positive}'', ``\textit{positive}'', ``\textit{neutral}'', ``\textit{negative}'', and ``\textit{very negative}''. To highlight the impact of calibration, we introduce a sixth, naive class label verbalized as:
\begin{quote}
    ``\textit{It was the best of times, it was the worst of times, it was the age of wisdom, it was the age of foolishness, it was the epoch of belief, it was the epoch of incredulity, it was the season of light, it was the season of darkness, it was the spring of hope, it was the winter of despair.}''
\end{quote}
taken from ``A Tale of Two Cities'' by Charles Dickens. This label is intentionally much longer than the other labels and consists of a well-known sequence of words, making it a strong candidate for unintended biases.

For simplicity, we use the Llama 3.1 (8B) model with a simple few-shot prompt:
\begin{quote}
    \small
    \ttfamily
    Text: This is a good product, and I'd probably recommend it to others. \\
    Label: positive \\[1ex]
    Text: The presentation was informative but not particularly engaging. \\
    Label: neutral \\[1ex]
    Text: The service was slow, and the food was cold. \\
    Label: negative \\[1ex]
    Text: This product has completely changed my life for the better! \\
    Label: 
\end{quote}

The model must predict the sentiment of the last sentence, whose ground-truth label is ``\textit{very positive}''. Table \ref{tab:calibration_example} presents the probabilities assigned to each class by the different methods.

\begin{table}
    \centering
    \setlength{\tabcolsep}{4pt}
    \small
    \begin{tabular}{lcccc}
        \toprule
        Class Label & Raw & Norm & CC & NCC \\
        \midrule
        very positive & 0.13 & 0.20 & 1e-9 & \textbf{0.81} \\
        positive & \textbf{0.86} & 0.39 & 2e-12 & 0.14 \\
        neutral & 2e-3 & 1e-3 & 7e-15 & 4e-4 \\
        negative & 3e-3 & 1e-3 & 3e-14 & 1e-3 \\
        very negative & 5e-5 & 4e-3 & 8e-12 & 9e-3 \\
        \textit{it was the best of times,} [...] & 1e-14 & \textbf{0.41} & \textbf{0.99} & 0.04 \\
        \bottomrule
    \end{tabular}
    \caption{Probabilities assigned to each class in the toy example. \textbf{Bold} values indicate the highest probability (i.e., the predicted class).}
    \label{tab:calibration_example}
\end{table}

The standard raw probability approach assigns a reasonable probability distribution but incorrectly predicts ``\textit{positive}'' instead of ``\textit{very positive}''. This happens because ``\textit{very positive}'' consists of two tokens (``\texttt{very}'' and ``\texttt{~positive}''), meaning its probability is the product of individual token probabilities, making it disproportionately lower.

Length normalization (Norm Prob) mitigates this length bias by normalizing multi-token probabilities. However, it assigns a high probability to the naive class label due to the high compound probability of its familiar token sequence, despite its irrelevance to the task.

CC (contextual calibration; \citealt{zhao2021calibrate}) attempts to correct label biases but fails in this case. Since the standard raw method assigns a very low prior probability to the naive (longest) label, CC overcompensates and amplifies its probability, leading to incorrect predictions.

NCC (normalized contextual calibration; \textit{ours}) successfully mitigates both length bias and overcompensation. By normalizing probabilities before calibration, it ensures that class probabilities are fairly adjusted, leading to a more accurate prediction of ``\textit{very positive}''.

This example highlights the importance of properly handling full-label probabilities, particularly when dealing with multi-token labels, to ensure reliable and unbiased predictions of LLMs.

\section{Dataset Details}
\label{app:dataset}

We evaluate NCC using eight text classification datasets and three multiple-choice QA datasets, spanning diverse domains and label complexities. Table \ref{tab:full_datasets} summarizes key characteristics including class count, balance status, test set size, and example labels.

\begin{table*}
    \centering
    \small
    \begin{tabularx}{\textwidth}{lcccX}
        \toprule
        Dataset & \# Classes & Balanced & $|\text{Test}|$ & Class Labels \\
        \midrule
        \multicolumn{5}{l}{\textit{Text Classification}}\\
        AG News \cite{agnews_dbpedia_yahoo_dataset} & 4 & \ding{51} & 500 & \quotes{\textit{World}}, \quotes{\textit{Sports}}, \quotes{\textit{Business}}, \quotes{\textit{Science and Technology}} \\
        SST-5 \cite{socher2013sst2_5} & 5 & \ding{55} & 1,100 & \quotes{\textit{very positive}}, \quotes{\textit{positive}}, \quotes{\textit{neutral}}, \quotes{\textit{negative}}, \quotes{\textit{very negative}} \\
        Yahoo \cite{agnews_dbpedia_yahoo_dataset} & 10 & \ding{51} & 500 & Topics like \quotes{\textit{Society}}, \quotes{\textit{Science}}, \quotes{\textit{Health}}, \quotes{\textit{Education}}, and others. \\
        DBpedia \cite{agnews_dbpedia_yahoo_dataset} & 14 & \ding{51} & 500 & Types of entities, such as \quotes{\textit{Company}}, \quotes{\textit{Educational Institution}}, \quotes{\textit{Artist}}, and more. \\
        20 Newsgroups \cite{20newsgroups_dataset} & 20 & \ding{55} & 500 & Categories are shown in Table \ref{tab:20newsgroups_mapping}. \\
        TREC-50 \cite{voorhees2000trec} & 50 & \ding{55} & 500 & Fine-grained question types like \quotes{\textit{Invention, book and other creative piece}}, \quotes{\textit{Description of a person}}, and \quotes{\textit{Distance, linear measure}}. \\
        Banking77 \cite{banking77_dataset} & 77 & \ding{55} & 924 & Intents such as \quotes{\textit{card activation}}, \quotes{\textit{transaction failed}}, \quotes{\textit{account blocked}}, and more. \\
        CLINC150 \cite{clinc150_dataset} & 150 & \ding{55} & 3,000 & Intents in domains like banking, travel, and work, such as \quotes{\textit{transfer money}}, \quotes{\textit{book flight}}, and \quotes{\textit{schedule meeting}}. \\
        \midrule
        \multicolumn{5}{l}{\textit{Multiple Choice QA}}\\
        OpenBookQA \cite{openbookqa_dataset} & 4 & N/A & 500 & N/A \\
        CommonsenseQA \cite{commonsenseqa_dataset} & 5 & N/A & 1,221 & N/A \\
        QASC \cite{qasc_dataset} & 8 & N/A & 926 & N/A \\
        \bottomrule
    \end{tabularx}
    \caption{Full dataset information.}
    \label{tab:full_datasets}
\end{table*}

\subsection{20 Newsgroups Dataset}
While widely adopted in traditional NLP research, the 20 Newsgroups dataset \citep{20newsgroups_dataset} has been largely overlooked in recent LLM-based classification studies due to its non-descriptive class labels (e.g., ``\textit{sci.med}''). In traditional methods, such as bag-of-words models, class label wording was irrelevant since these approaches relied solely on feature representations of the input text. However, LLMs benefit from more representative and meaningful labels, as they utilize the wording of class labels during inference. To address this, we remap the original categories as shown in Table \ref{tab:20newsgroups_mapping} to enhance interpretability and facilitate a more effective use of this dataset with LLMs.

\begin{table}[h]
    \centering
    \small
    \begin{tabular}{l|l}
        \toprule
        \textbf{Original Class Label} & \textbf{New Class Label} \\ \midrule
        alt.atheism & Atheism \& Secularism \\
        comp.graphics & Computer Graphics \\
        comp.os.ms-windows.misc & Windows OS \\
        comp.sys.ibm.pc.hardware & IBM PC Hardware \\
        comp.sys.mac.hardware & Macintosh Hardware \\
        comp.windows.x & X Window System \\
        misc.forsale & Classified Ads \\
        rec.autos & Automobiles \\
        rec.motorcycles & Motorcycles \\
        rec.sport.baseball & Baseball \\
        rec.sport.hockey & Hockey \\
        sci.crypt & Cryptography \\
        sci.electronics & Electronics \\
        sci.med & Medicine \\
        sci.space & Space \& Astronomy \\
        soc.religion.christian & Christianity \\
        talk.politics.guns & Gun Politics \\
        talk.politics.mideast & Middle East Politics \\
        talk.politics.misc & General Politics \\
        talk.religion.misc & General Religion \\
        \bottomrule
    \end{tabular}
    \caption{Mapping of the original 20 Newsgroups class labels to more representative and descriptive labels for use with LLMs.}
    \label{tab:20newsgroups_mapping}
\end{table}

\section{Detailed Results}
\label{app:results}

\subsection{Model Size Comparison}
\label{app:model_size}

In the main text, we demonstrate the effectiveness of NCC across different LLMs of different sizes (Section \ref{sec:results}). To further evaluate the impact of model size on NCC performance, we stick with the Llama 3.1 model and compare its performance with different sizes (8B, 70B) on the text classification task. 

\begin{figure}
    \centering
    \includegraphics[width=\linewidth]{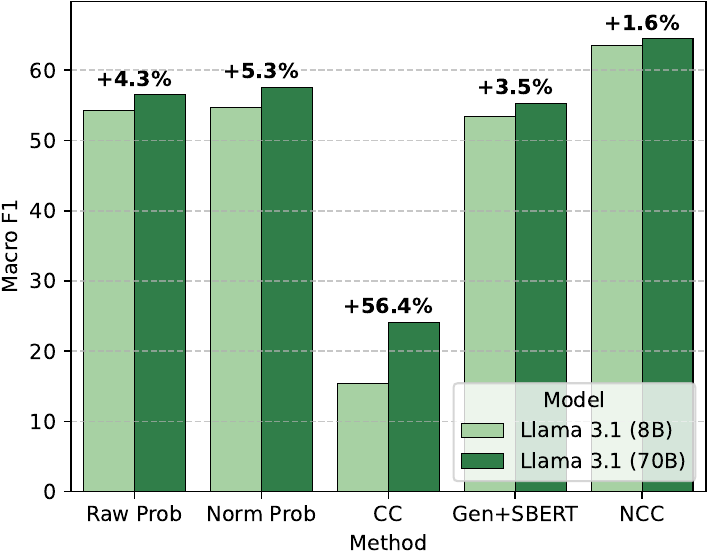}
    \caption{Average performance of all methods with Llama 3.1 (8B) and Llama 3.1 (70B). Numbers indicate relative improvements when scaling from 8B to 70B.}
    \label{fig:plot_by_method_and_model}
\end{figure}

Figure \ref{fig:plot_by_method_and_model} shows the average performance and relative improvement from the 8B version to the 70B version of Llama 3.1 for each method under the same settings. NCC enables the smaller model to achieve the most similar performance to the larger model, with only a 1.6\% gap in average F1, despite being almost nine times smaller. This is a significant drop in performance gap, compared to the 4.3\% difference across model sizes in the standard raw probability approach. This result highlights the effectiveness of NCC in enhancing the performance of smaller models, enabling them to achieve comparable results to larger models in text classification when properly calibrated.

\subsection{Zero-Shot Results}
\label{app:zero_shot}

Table \ref{tab:results_k0} provides the full results for all methods on each dataset and model in the zero-shot setting (corresponding to Figure \ref{fig:plot_by_model_and_method_0} in the main text).

\begin{table*}[h]
    \small
    \centering
    \setlength{\tabcolsep}{4pt}
    \begin{tabular}{lcccccccc|c}
    \toprule
    Method & AG News & SST-5 & Yahoo & DBpedia & 20 Newsgroups & TREC-50 & Banking77 & CLINC150 & Avg. \\
    \midrule
    \multicolumn{10}{c}{\textit{Llama 3.1 (8B)}} \\
    \midrule
    Raw Prob & $\mathbf{58.5}$ & $20.6$ & $16.2$ & $36.8$ & $40.8$ & $12.5$ & $34.2$ & $25.4$ & $30.6$ \\
    Norm Prob & $17.7$ & $14.0$ & $36.0$ & $57.6$ & $18.7$ & $3.5$ & $26.5$ & $25.4$ & $24.9$ \\
    CC & $10.0$ & $16.4$ & $11.6$ & $1.0$ & $5.7$ & $0.2$ & $0.0$ & $0.7$ & $5.7$ \\
    Gen+SBERT & $56.2$ & $\mathbf{24.0}$ & $45.9$ & $44.7$ & $29.1$ & $\mathbf{20.2}$ & $\mathbf{47.2}$ & $\mathbf{55.9}$ & $40.4$ \\
    \cmidrule(lr){2-10}
    NCC & $57.8$ & $23.3$ & $\mathbf{52.6}$ & $\mathbf{67.4}$ & $\mathbf{57.8}$ & $19.9$ & $41.5$ & $53.3$ & $\mathbf{46.7}$ \\
    \midrule
    \multicolumn{10}{c}{\textit{Mistral 7B}} \\
    \midrule
    Raw Prob & $\mathbf{54.8}$ & $14.0$ & $10.3$ & $30.2$ & $39.5$ & $12.9$ & $\mathbf{31.6}$ & $14.5$ & $26.0$ \\
    Norm Prob & $31.6$ & $10.8$ & $13.3$ & $5.2$ & $23.1$ & $5.6$ & $27.8$ & $16.8$ & $16.8$ \\
    CC & $10.0$ & $15.8$ & $11.0$ & $1.0$ & $0.4$ & $0.2$ & $0.0$ & $0.5$ & $4.9$ \\
    Gen+SBERT & $43.5$ & $21.9$ & $26.3$ & $34.7$ & $6.6$ & $13.5$ & $25.2$ & $18.1$ & $23.7$ \\
    \cmidrule(lr){2-10}
    NCC & $48.9$ & $\mathbf{26.4}$ & $\mathbf{41.9}$ & $\mathbf{44.0}$ & $\mathbf{48.9}$ & $\mathbf{15.9}$ & $30.5$ & $\mathbf{36.6}$ & $\mathbf{36.7}$ \\
    \midrule
    \multicolumn{10}{c}{\textit{Qwen2.5 (7B)}} \\
    \midrule
    Raw Prob & $54.3$ & $22.7$ & $18.3$ & $49.2$ & $32.9$ & $20.3$ & $29.0$ & $14.5$ & $30.2$ \\
    Norm Prob & $46.5$ & $\mathbf{31.1}$ & $34.0$ & $65.9$ & $22.2$ & $7.6$ & $32.1$ & $28.5$ & $33.5$ \\
    CC & $10.0$ & $12.9$ & $20.7$ & $1.0$ & $1.3$ & $0.2$ & $0.0$ & $0.6$ & $5.8$ \\
    Gen+SBERT & $38.0$ & $16.2$ & $31.6$ & $38.8$ & $19.5$ & $19.2$ & $36.4$ & $33.1$ & $29.1$ \\
    \cmidrule(lr){2-10}
    NCC & $\mathbf{57.7}$ & $23.5$ & $\mathbf{45.3}$ & $\mathbf{69.1}$ & $\mathbf{50.3}$ & $\mathbf{29.2}$ & $\mathbf{36.5}$ & $\mathbf{40.8}$ & $\mathbf{44.1}$ \\
    \midrule
    \multicolumn{10}{c}{\textit{GPT-J (6B)}} \\
    \midrule
    Raw Prob & $53.3$ & $16.9$ & $8.3$ & $42.4$ & $34.1$ & $15.8$ & $27.8$ & $26.1$ & $28.1$ \\
    Norm Prob & $26.2$ & $13.8$ & $14.2$ & $52.2$ & $17.2$ & $15.2$ & $24.1$ & $22.8$ & $23.2$ \\
    CC & $10.0$ & $10.6$ & $1.8$ & $1.0$ & $3.0$ & $0.2$ & $0.0$ & $0.9$ & $3.4$ \\
    Gen+SBERT & $53.3$ & $\mathbf{21.3}$ & $38.7$ & $47.5$ & $18.7$ & $19.1$ & $\mathbf{37.9}$ & $38.5$ & $34.4$ \\
    \cmidrule(lr){2-10}
    NCC & $\mathbf{59.8}$ & $21.0$ & $\mathbf{47.0}$ & $\mathbf{52.8}$ & $\mathbf{45.8}$ & $\mathbf{25.0}$ & $37.4$ & $\mathbf{50.4}$ & $\mathbf{42.4}$ \\
    \midrule
    \multicolumn{10}{c}{\textit{Llama 3.1 (70B)}} \\
    \midrule
    Raw Prob & $57.5$ & $22.2$ & $15.9$ & $40.3$ & $43.1$ & $14.9$ & $33.5$ & $28.7$ & $32.0$ \\
    Norm Prob & $22.1$ & $30.2$ & $37.4$ & $56.4$ & $26.0$ & $13.4$ & $22.9$ & $37.8$ & $30.8$ \\
    CC & $10.0$ & $16.2$ & $22.2$ & $1.0$ & $5.7$ & $0.2$ & $0.0$ & $0.6$ & $7.0$ \\
    Gen+SBERT & $54.8$ & $24.0$ & $44.5$ & $45.1$ & $28.4$ & $18.5$ & $\mathbf{39.1}$ & $51.6$ & $38.3$ \\
    \cmidrule(lr){2-10}
    NCC & $\mathbf{64.3}$ & $\mathbf{32.5}$ & $\mathbf{51.9}$ & $\mathbf{62.2}$ & $\mathbf{58.0}$ & $\mathbf{20.5}$ & $38.0$ & $\mathbf{53.3}$ & $\mathbf{47.6}$ \\
    \bottomrule
    \end{tabular}
    \caption{Performance comparisons (macro-F1) for the zero-shot setting. There is no ${\pm\text{std}}$ because there are no examples, resulting in consistent performance across runs.}
    \label{tab:results_k0}
\end{table*}

\subsection{Comparison with Other Calibration Methods}
\label{app:other_calibration_methods}

In the main text, we compare NCC with contextual calibration (CC; \citealt{zhao2021calibrate}) as a representative standard calibration method, due to its simplicity and wide recognition in the community.
However, there are several other relevant calibration methods proposed in recent literature.

\begin{table*}
    \small
    \centering
    \setlength{\tabcolsep}{4pt}
    \begin{tabular}{lcccccccc|c}
    \toprule
    Method & AG News & SST-5 & Yahoo & DBpedia & 20 Newsgroups & TREC-50 & Banking77 & CLINC150 & Avg. \\
    \midrule
    \multicolumn{10}{c}{\textit{No Calibration}} \\
    \midrule
    Raw Prob & 83.8 & \textbf{42.3} & 61.3 & 83.4 & 41.0 & 28.2 & 37.8 & 55.9 & 54.2 \\
    Norm Prob & 85.3 & 38.6 & 60.6 & 80.0 & 43.3 & 29.3 & 36.3 & 64.1 & 54.7 \\
    Gen+SBERT & 80.4 & 41.2 & 64.3 & 67.3 & 40.5 & \textbf{33.8} & 47.5 & 52.3 & 53.4 \\
    \midrule
    \multicolumn{10}{c}{\textit{Standard Calibration}} \\
    \midrule
    CC & 84.0 & 23.9 & 8.1 & 1.0 & 5.3 & 0.2 & 0.1 & 1.0 & 15.4 \\
    DC & 80.6 & 23.5 & 25.2 & 1.0 & 9.0 & 0.2 & 0.1 & 2.1 & 17.7 \\
    GC & 86.1 & 18.8 & 47.4 & 15.2 & 6.5 & 0.2 & 0.1 & 3.3 & 22.2 \\
    BC & \textbf{86.9} & 33.2 & 49.9 & 1.0 & 0.4 & 0.2 & 2.8 & 3.4 & 22.2 \\
    \midrule
    \multicolumn{10}{c}{\textit{Normalized Calibration}} \\
    \midrule
    NCC & 78.0 & 38.2 & \textbf{65.1} & \textbf{92.5} & 67.7 & \textbf{33.8} & 59.5 & 73.1 & 63.5 \\
    NDC & 83.8 & \textbf{44.2} & 63.6 & 92.3 & 68.8 & 32.2 & \textbf{64.4} & \textbf{75.5} & 65.6 \\
    NGC & 83.6 & 44.6 & 61.2 & 90.8 & 67.7 & 22.4 & 56.9 & 69.2 & 62.1 \\
    NBC & 83.8 & \textbf{49.5} & 64.9 & 92.0 & \textbf{69.8} & 32.8 & 61.1 & 73.4 & \textbf{65.9} \\
    \bottomrule
    \end{tabular}
    \caption{Comparison of different calibration methods with Llama 3.1 (8B). CC = contextual calibration \citep{zhao2021calibrate}; DC = domain-context calibration \citep{fei2023domaincalibration}; GC = generative calibration \citep{jiang2023generative}; BC = batch calibration \citep{zhou2024batchcalibration}; NCC = normalized contextual calibration; NDC = normalized domain-context calibration; NGC = normalized generative calibration; NBC = normalized batch calibration.}
    \label{tab:other_calibration_methods}
\end{table*}

Here, we extend the comparison to include domain-context calibration (DC; \citealt{fei2023domaincalibration}), generative calibration (GC; \citealt{jiang2023generative}), and batch calibration (BC; \citealt{zhou2024batchcalibration}).
For these methods, we follow the original papers' implementations (e.g., using random contexts from the training set for DC, and L = 100 generated inputs from the model for GC to estimate priors).
Moreover, we apply our normalized calibration approach to these methods, resulting in normalized domain-context calibration (NDC), normalized generative calibration (NGC), and normalized batch calibration (NBC).

Given the computational cost of evaluating all methods across multiple models and datasets, we limit this comparison to the Llama 3.1 (8B) model on the text classification datasets.

The results are shown in Table \ref{tab:other_calibration_methods}. Overall, for standard calibration, DC, GC, and BC generally improve over CC on datasets where label length bias is limited (e.g., AG News, SST-5, Yahoo), but they still struggle when multi-token label bias is strong (e.g., DBpedia, 20 Newsgroups, TREC-50, Banking77, CLINC150), where NCC consistently performs best.
For normalized calibration, augmenting these methods with length normalization (NDC/NGC/NBC) can yield further gains over NCC, but these improvements are smaller than the gaps between NCC and CC/DC/GC/BC, indicating that the main performance boost comes from combining some form of semantic calibration with length normalization.

\section{Reliability of the LLMs Confidences}
\label{app:reliability}

As shown in Section~\ref{sec:analysis}, NCC improves the reliability of model confidence scores. Here, we extend that analysis by examining how often labels are predicted based on their length (left histograms) and how prediction confidences are distributed (right histograms). We also report the enrichment factor, which quantifies how much more frequently certain labels are predicted compared to the ground truth distribution.
Label lengths are normalized within each dataset, such that the shortest label corresponds to 0\%, the longest to 100\%, and intermediate labels are scaled proportionally.
This analysis is conducted using the Llama 3.1 (8B) model across all classification datasets from the main text.

\begin{figure}
    \centering
    \includegraphics[width=\linewidth]{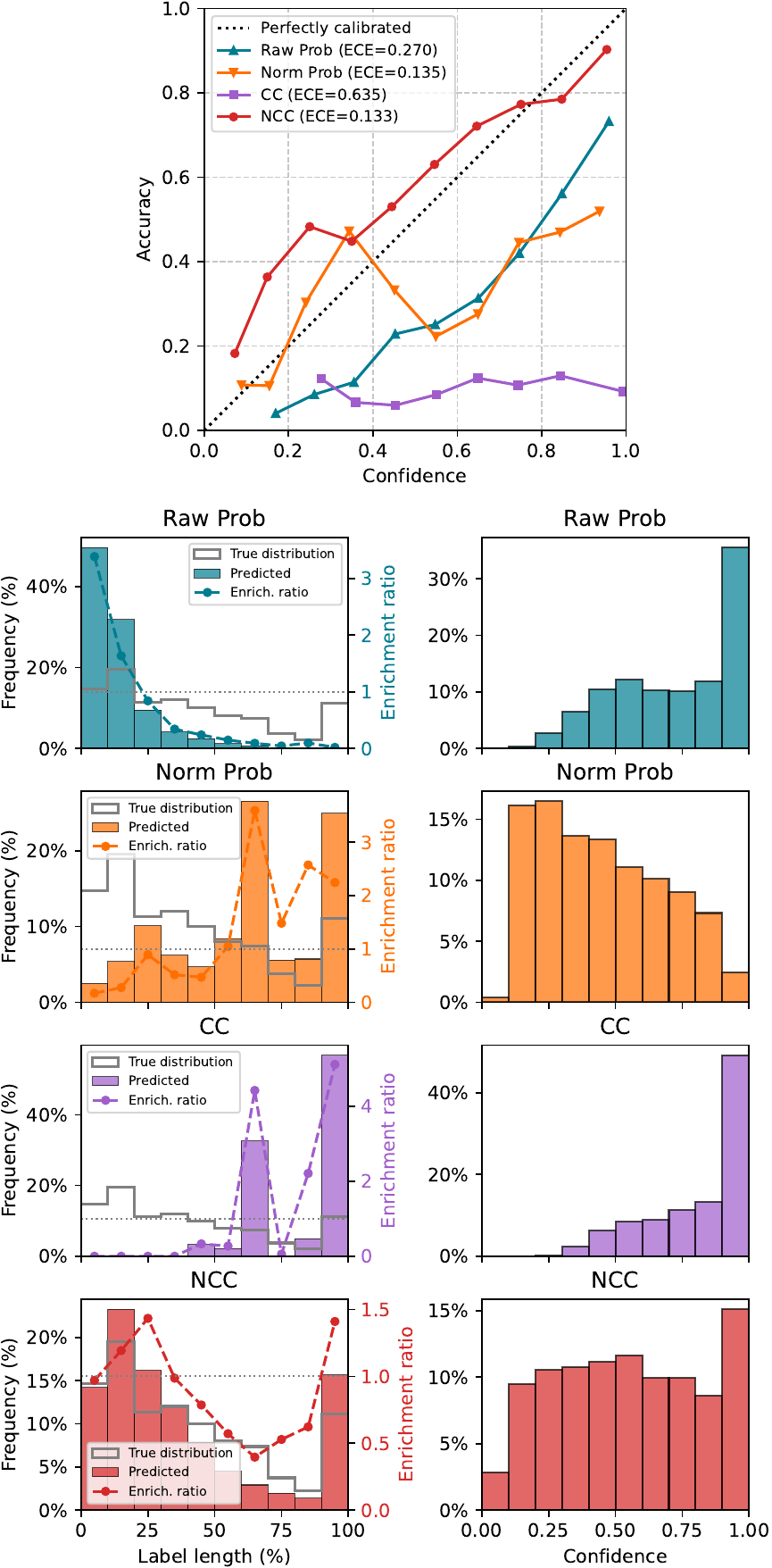}
    \caption{Reliability diagrams and frequency of label prediction by label length (left) and confidence (right).}
    \label{fig:reliability_full}
\end{figure}

The results, summarized in Figure~\ref{fig:reliability_full}, further highlight the label length bias problem: the raw probabilities from the LLM strongly favor shorter labels, with an enrichment factor exceeding 3, and predicts them with high confidence (near certainty in over 30\% of cases).

Length normalization (Norm Prob) reduces this bias by flattening probabilities across labels of different lengths, producing a confidence distribution skewed toward lower values (i.e., most predictions fall within the 10\%--50\% confidence range).
However, as discussed in Section~\ref{sec:length_bias}, it introduces a new issue: later tokens in longer labels become disproportionately likely, resulting in an over-prediction of long labels.
Similarly, standard CC (without length normalization) overcompensates for the length bias, favoring longer labels instead, as analyzed in Appendix~\ref{app:calibration}.

In contrast, NCC produces label predictions that closely match the true distribution across label lengths, effectively mitigating both short- and long-label biases. Additionally, its prediction confidence is more evenly distributed and more closely resembles a well-calibrated model, reflected by the lowest ECE in the top plot of Figure~\ref{fig:reliability_full}.

\section{Multi-Token vs.\ Single-Token Labels}
In the main text of the paper, we demonstrate improved label bias mitigation by applying NCC, a calibration technique at the full-label level to solve the identified label length bias when working with multi-token labels.
Previous works do not consider this type of bias, as they focus on single-token labels by remapping original multi-token labels to single-token alternatives \cite{zhao2021calibrate,fei2023domaincalibration,han2023prototypicalcalibration,jiang2023generative,zhou2024batchcalibration}.
For instance, in sentiment analysis tasks, labels like \quotes{\textit{very positive}} and \quotes{\textit{very negative}} are often replaced by \quotes{\textit{great}} and \quotes{\textit{terrible}}, respectively.
These label remappings are not always feasible, and it is ambiguous to determine when a label can be represented by a single token without the risk of losing semantic meaning. For example, in the Banking77 dataset \cite{banking77_dataset} labels like \quotes{\textit{balance not updated after cheque or cash deposit}} and \quotes{\textit{top up failed}} contain nuances hardly distinguishable by means of an alternative single-token label.

However, one question that remains open is whether, when possible, considering single-token alternatives yields better results than the original multi-token labels. To address this, we conduct an additional analysis comparing the performance of the original multi-token labels against the widely adopted remappings of the SST-5 dataset \cite{socher2013sst2_5}.
We consider the original labels and the single-token alternatives presented in Table \ref{tab:sst5_mapping}.
We run this analysis with the Llama 3.1 (8B) model in both zero- and few-shot settings. As in our main methodology, the few-shot setting is run with 5 different random seeds.

\begin{table}
    \centering
    \small
    \begin{tabular}{l|l}
        \toprule
        \textbf{Multi-token (Original)} & \textbf{Single-token (Remapping)} \\ \midrule
        very positive & great \\
        positive & good \\
        neutral & neutral \\
        negative & bad \\
        very negative & terrible \\
        \bottomrule
    \end{tabular}
    \caption{Mapping of the original SST-5 multi-token class labels to single token alternatives, as done in prior work.}
    \label{tab:sst5_mapping}
\end{table}

\begin{figure}
    \centering
    \begin{subfigure}{\columnwidth}
        \centering
        \includegraphics[width=0.8\textwidth]{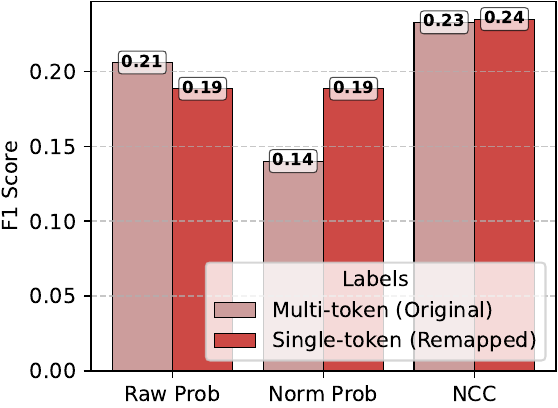}
        \caption{Zero-shot ($k=0$).}
        \label{subfig:single_vs_multi_0}
    \end{subfigure}

    \vspace{1em}
    
    \begin{subfigure}{\columnwidth}
        \centering
        \includegraphics[width=0.8\textwidth]{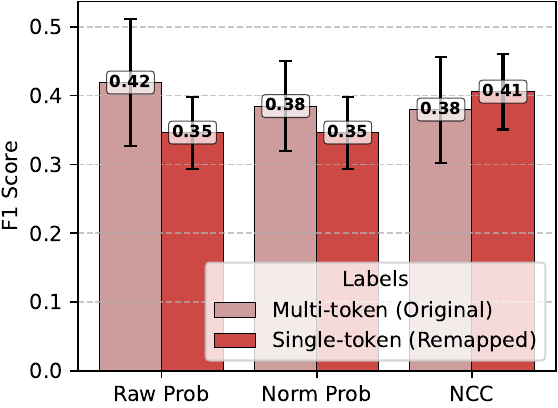}
        \caption{Few-shot ($k=5$).}
        \label{subfig:single_vs_multi_5}
    \end{subfigure}
    
    \caption{Performance comparison of original multi-token labels against single-token alternatives.}
    \label{fig:single_vs_multi}
\end{figure}

Figure~\ref{fig:single_vs_multi} compares the standard raw probability approach, probability with length normalization, and NCC when using either the original multi-token labels or simplified single-token alternatives.
In the standard Raw Prob setting, multi-token labels are slightly favored, though the difference remains within the error margins.
A notable discrepancy emerges in the \emph{zero-shot} setting with the Norm Prob method, where multi-token labels containing \quotes{\textit{very X}}\footnote{Where \quotes{\textit{X}} denotes either \quotes{\textit{positive}} or \quotes{\textit{negative}}.} are over-predicted. This occurs because the token \quotes{\textit{X}} following \quotes{\textit{very}} receives higher conditional probability than when predicted alone, leading to inflated overall label scores -- a problem also discussed in Section~\ref{sec:length_bias}.
In contrast, NCC performs consistently well across both label types, demonstrating its ability to mitigate this bias.
These results indicate that retaining the original multi-token labels does not hinder performance and avoids potential semantic loss introduced by label simplification.

\end{document}